\documentclass[runningheads]{llncs}
\usepackage{graphicx}
\usepackage{subfig}
\usepackage{booktabs}
\usepackage{comment}
\usepackage{amsmath,amssymb} 
\usepackage{color}

\usepackage[width=122mm,left=12mm,paperwidth=146mm,height=193mm,top=12mm,paperheight=217mm]{geometry}
\usepackage{hyperref}

\begin{document}
\pagestyle{headings}
\mainmatter
\def\ECCVSubNumber{2769}  

\title{Training End-to-end Single Image Generators without GANs} 

\titlerunning{Training End-to-end Single Image Generators without GANs}
%
\author{Yael Vinker \and
Nir Zabari \and
Yedid Hoshen}
\authorrunning{Y. Vinker et al.}
%
\institute{School of Computer Science and Engineering \\ The Hebrew University of Jerusalem, Israel.}
\maketitle

\begin{abstract}
We present AugurOne, a novel approach for training single image generative models. Our approach trains an upscaling neural network using non-affine augmentations of the (single) input image, particularly including non-rigid thin plate spline image warps. The extensive augmentations significantly increase the in-sample distribution for the upsampling network enabling the upscaling of highly variable inputs. A compact latent space is jointly learned allowing for controlled image synthesis.  Differently from Single Image GAN, our approach does not require GAN training and takes place in an end-to-end fashion allowing fast and stable training. We experimentally evaluate our method and show that it obtains compelling novel animations of single-image, as well as, state-of-the-art performance on conditional generation tasks e.g. paint-to-image and edges-to-image. Image animation results are presented on our \href{http://www.vision.huji.ac.il/augurone/}{\textcolor{blue}{project page}}.
\end{abstract}

\section{Introduction}
\label{sec:intro}

Learning image distributions is a key computer vision task with many applications such as novel image synthesis, image priors and image translation. Many methods were devised for learning image distributions either by directly learning the image probability density function or by aligning the distribution of images with a simpler parametric distribution. Current methods normally require large image collections for learning faithful models of image distributions. This limits the applicability of current methods to images which are similar to many other images (such as facial images) but not to the long-tail of unique images. Even when large image collections exist, the task of learning image generators remains very difficult for complex images i.e. learning models for large image collections such as ImageNet or Places365 is still the subject of much research. 

\begin{figure}[t]
\begin{center}
\includegraphics[width=1.0\linewidth]{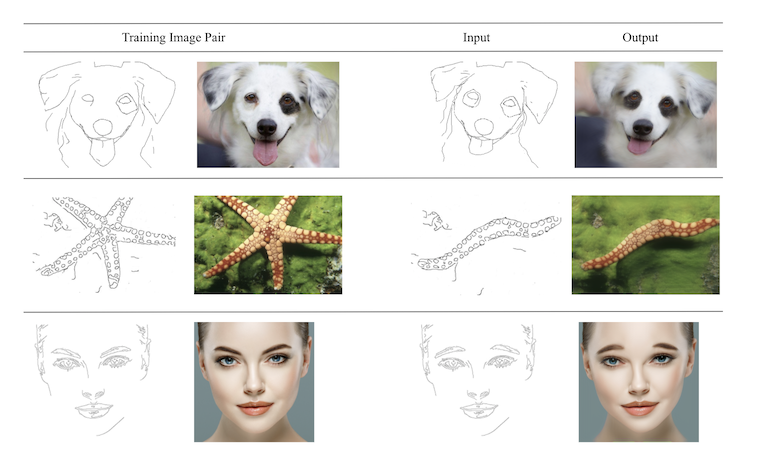}
\end{center}
\caption{Edges2Images: Our method takes a single training pair of edges and image. After training, we input a new edge map, our model synthesizes a novel image with the correct edges.}
\end{figure}

A complementary approach to learning image generators from large collections is learning a model for each image. Although the sample size is far smaller, a single image has a simpler distribution which may be simpler to learn. Very recently, SinGAN, a general purpose single image generative adversarial network model was proposed by Shaham et al. \cite{shaham2019singan}. This pioneering work, utilizes the multiple patch pyramids present in a single image to train a cascade of simple conditional adversarial networks. It was shown to be able to perform a set of conditional and unconditional image synthesis task. Despite the amazing progress achieved by this work, SinGAN has several drawbacks mainly due to the the use of generative adversarial (GAN) training. The stability issues inherent in GAN training, motivated the architectural choice of using a cascade of generators rather than a single end-to-end network, which is more brittle and less convenient. Similarly to other PatchGAN-based methods, it can only deal with texture-like image details rather than large objects.    

In recent years, an alternative set of techniques to generative adversarial networks has been developed. These techniques, sometimes named "non-adversarial" methods, attempt to perform image generative modeling without the use of adversarial training. non-adversarial methods have achieved strong performance on unconditional image generation (VAE \cite{kingma2014auto}, GLO \cite{glo}, GLANN \cite{hoshen2019non}, VQ-VAE2 \cite{razavi2019generating}), as well as unsupervised image translation tasks (\cite{nam_eccv}, \cite{cohen2019bidirectional}). In this work, we extend non-adversarial training to single-image unconditional image generation tasks as well as a set of conditional generation tasks.

To contextualize SinGAN within the process of non-adversarial learning, we offer an insightful interpretation for the operation of SinGAN. Our main insight is that SinGAN is a combination of a super-resolution network and GAN-based image augmentations of the single input image. All these networks are learned in a stage-wise rather than end-to-end manner. The cascaded training is due to the difficulty of GAN training. As insufficient variation is present within a single image, augmentation of the training image is necessary. 

We propose an alternative non-adversarial approach, AugurOne, which takes the form of an upsampling network. AugurOne learns to upsample a downscaled version of the input image by reconstructing it at high resolution (much like single image super-resolution). As a single image does not present sufficient variation, we augment the training image with carefully crafted affine and non-affine augmentations. AugurOne is trained end-to-end with a non-adversarial perceptual reconstruction loss. In cases that require synthesis of novel image samples, we add a front-end variational autoencoder which learns a compact latent space allowing novel image generation by latent interpolation. Novel images of arbitrary size are synthesized using an interpolations between concatenations of different augmentations of the input images. Our method enjoys fast and stable training and achieves very strong results on novel image synthesis, most remarkable for image animation with large object movements. This is enabled by our encoder allowing control over the novel synthesized images. Our method achieves very compelling results on conditional generation tasks e.g. paint-to-image and edges-to-image. 

\section{Previous Work}
\label{sec:prev}

\textbf{Generative Modeling:} Image generation has attracted research for several decades. Some successful early approach used Gaussian or mixtures of Gaussians models (GMM) \cite{zoran2011learning}. Due to the limited expressive power of GMM, such methods achieved limited image resolutions and quality and mainly focused on modeling image patches rather than large images. Over the last decade, with the advent of deep neural network models and increasing dataset sizes, significant progress was made on image generation models. Early deep models include Reduced Boltzmann Machines (RBMs). Variational Autoencoders \cite{vae}, first introduced by Kingma and Welling made a significant breakthrough as a principled model for mapping complex empirical distributions (e.g. images) to simple parametric distributions (e.g. Gaussian). Although VAEs are relatively simple to train and have solid theoretical foundations, the images they generate are not as sharp as those generated by other state-of-the-art methods. Auto-regressive and flow-based models \cite{dinh2014nice} \cite{kingma2018glow} \cite{oord2016pixel}\cite{oord2016wavenet} have also been proposed as a promising direction for image generative models.

\textbf{Adversarial Generative Models:} Currently, the most popular paradigm for training image generation models is Generative Adversarial Networks (GANs). GANs were first introduced by Goodfellow et al. \cite{goodfellow2014generative} and are currently used in computer vision for three main uses: i) unconditional image generator training \cite{dcgan} ii) unsupervised image translation between domains \cite{CycleGAN2017} \cite{discogan} iii) serving as a perceptual image loss function \cite{pix2pix}. GANs are able to synthesize very sharp images but suffer from some notable drawbacks, particularly very sensitive training and mode dropping (hurting generation diversity). Overcoming the above limitations has been the focus of much research over the last several years. One mitigation is changing the loss function to prevent saturation (e.g. Wasserstein GAN \cite{arjovsky2017wasserstein}). Another mitigation is using different types of discriminator regularizations (where the aim is typically Lipschitzness). Regularization methods include: clipping \cite{arjovsky2017wasserstein}, gradient regularization \cite{gulrajani2017improved} \cite{Mescheder2018ICML} or spectral normalization \cite{miyato2018spectral}. 

\textbf{Non-Adversarial Methods:} An alternative direction motivated by the limitations of GAN methods is the development of non-adversarial methods for image generation. Some notable methods include: GLO \cite{glo} and IMLE \cite{imle}. Hoshen at al.~\cite{hoshen2019non} combine GLO and IMLE into a new method, GLANN which is able to synthesize sharp images from a parametric distribution. It was able to outperform GANs in on a low-resolution benchmark. VQ-VAE2 \cite{razavi2019generating} also consists of a two-step approach, a combination of a vector-quantization VAE and an auto-regressive pixelCNN model and was able to achieve very high resolution image generation competitive with state-of-the-art GANs. Non-adversarial methods have also been successfully introduced for supervised image-mapping (Chen and Koltun \cite{chen2017photographic}), and unsupervised disentanglement (LORD \cite{gabbay2019demystifying}). In this work, we present a non-adversarial alternative for training unconditional image generation from a single-image.  

\textbf{Single image generators:} Limited work was done on training image generators from a single-image due to difficulty of the task. Deep Image Prior is a notable work which shows that training a deep network on a single image can form an effective image prior, however this work cannot perform unconditional image generation. Previous work was also performed on training image inpainting \cite{ulyanov2018deep} and super-resolution \cite{shocher2018zero} from a single image however these works are limited to a single application and perform conditional generation. Our work draws much inspiration from the seminal work of Shaham et al. \cite{shaham2019singan}, and presents several novelties which are demonstrated to be advantageous. Our method is non-adversarial and therefore enjoys fast, stable and end-to-end training. It uses augmentations that are explainable (as opposed the the more opaque GAN) giving control over the learning process. Our method is also able to deal with larger-scale objects leading to attractive animations from a single image, as well as unsupervised domain translation between paint to photo-realistic images. It is also very effective for conditional generation e.g. mapping edges to image given a single training pair.

\section{Analysis of SinGAN}
\label{sec:singan}

In this section, we analyze SinGAN from the perspective of non-adversarial learning. 

SinGAN consists of a cascade of upscaling networks. Each network takes as an input an upscaled low-res image, combined with noise and learns to predict the residuals between the upscaled low-res image and the high-res image. The generators are trained sequentially on the output of the previous (low-resolution) generator.

\begin{equation}
    \label{eq:singan_upsample}
    x^{HR} = x^{LR} + G_n(x^{LR} + noise)
\end{equation}

Additionally, a noise to low-res image GAN $G_0(noise)$ is learned at the lowest resolution level. This unconditional generator learns to take in a low-resolution image, where each pixel was sampled from a random normal distribution. 

\begin{equation}
    \label{eq:singan_g0}
    x_{random}^{LR} = G_0(noise)
\end{equation}

There are two losses used: i) an $L_2$ reconstruction loss for $G_n$ where $n > 0$ - this is a standard super-resolution loss ii) an adversarial loss at every level.

\begin{figure}[t]
\begin{center}
\begin{tabular}{@{\hskip0pt}c@{\hskip4pt}c@{\hskip2pt}c@{\hskip2pt}c@{\hskip2pt}c@{\hskip2pt}c@{\hskip2pt}c@{\hskip2pt}c}

Training Image & Sample 1 & Sample 2 & Sample 3 \\
\midrule
\includegraphics[width=0.2\linewidth]{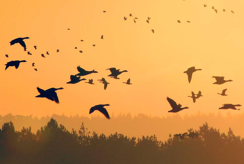} &
\includegraphics[width=0.2\linewidth]{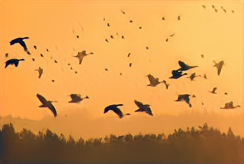} &
\includegraphics[width=0.2\linewidth]{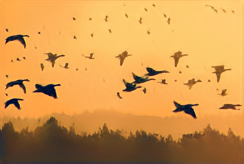} &
\includegraphics[width=0.2\linewidth]{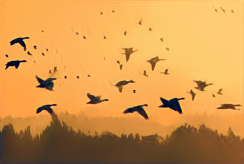} \\
\vspace{0.5em}\\
Training Image & Input & With noise & No noise \\
\midrule
\includegraphics[width=0.2\linewidth]{figures/paint2image/birds.png} &
\includegraphics[width=0.2\linewidth]{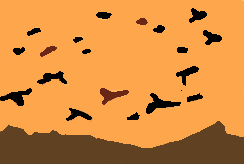} &
\includegraphics[width=0.2\linewidth]{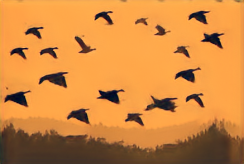} &
\includegraphics[width=0.2\linewidth]{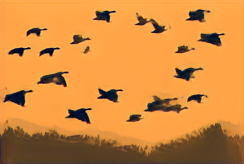} \\
\end{tabular}

\end{center}
\caption{(top) Several novel image samples generated by SinGAN when trained without noise (bottom) A comparison of paint2image transfer generated by SinGAN when trained with and without noise. In both cases the results are comparable suggesting that noise may be omitted from single image generator training.}  
\label{fig:noise_exp}
\end{figure}

In order to construct a non-adversarial alternative to SinGAN, we reinterpret this method as consisting of two sets of familiar operations. First, a set of super-resolution networks $G_n,n>0$ trained using a standard combination of PatchGAN and $L_2$ losses as a perceptual loss. This is a standard conditional generation approach, also used for super-resolution. It suffers from two drawbacks on its own; a single image is not sufficient for training due to overfitting and it does not supply the means for generating novel images. SinGAN therefore combines it with a low-resolution GAN which serves two purposes: it allows for generating novel images and it provides augmentation for the single training image, allowing for training deeper networks. Additionally, we hypothesize the noise in upscaling generators ($G_n, n>0$) is typically not a critical component for training SinGANs.   

To validate our analysis of SinGAN, we performed a supporting experiment. We removed the noise from the training of all the conditional generators i.e. from all $G_n,n>0$ but not from $G_0$. We can see random samples in Fig.~\ref{fig:noise_exp}. The random generation is of similar quality to that obtained with noise. 

To conclude, for training a single image generator having the capability of SinGAN, there are several requirements i) a perceptual loss for evaluating the upscaling ability of generators $G_n$ ii) a method for augmenting the single input image $x$ iii) a principled method for generating novel images. A conditional single image generator has only the first two requirements.  In Sec.~\ref{sec:method}, we will present a novel non-adversarial method, AugurOne, for training single image generators.

\section{Method}
\label{sec:method}

In this section, we propose a principled non-adversarial method for training single image generators. Our method is trained end-to-end and is fast and robust.   

\subsection{End-to-end image upscaling}
\label{sec:method:upscaling}

One of the conclusions of Sec.~\ref{sec:singan} is that the basic component in a single-image generator training is a high-quality upscaling network. Such a network is already sufficient for single-image conditional tasks such as harmonization, Edges2Image or Paint2Image. There were two main challenges solved by adversarial methods: the small sample size (just a single image) and an effective perceptual loss.

We propose a non-adversarial solution for training an upscaling network. Instead of using a cascade of generators, we simply train a single multi-scale generator network. The generator architecture is identical to that of SinGAN, however it is trained end-to-end. The input is an image at the lowest resolution, the expected outputs are images of the entire set of scales.

\begin{equation}
    \label{eq:upscaling_e2e}
    x^{1},x^{2},...,x^{N} = G(x_{actual}^{0})
\end{equation}

The loss function consists of the sum of reconstruction errors between the predicted and actual images across the entire image pyramid.

\begin{equation}
    \label{eq:upscaling_loss}
    L_{upscaling}(G) = \sum_n \ell(x^{n}, x_{actual}^{n})
\end{equation}

\begin{figure}[t]
\begin{center}
\includegraphics[width=1\linewidth]{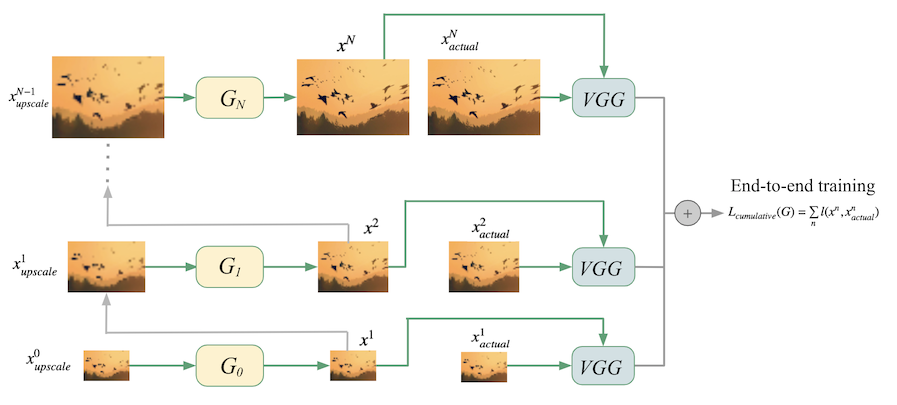}
\end{center}
\caption{An illustration of our upsampling network. The network takes as input the lowest resolution image $x^0_{actual}$. Each block $G_i$ takes as input the bi-cubically upscaled low-resolution image and learns the residual so that it reconstruct the groundtruth higher-resolution image $x^i_{actual}$. Our network trains all blocks end-to-end, each up-scaling the input image by a constant scale factor. Differently from SinGAN that trains each block separately, we use non-adversarial learning allowing us to train the entire network end-to-end by summing the reconstruction losses from all the levels. For more perceptually pleasing results, we use a VGG perceptual loss.}  
\label{fig:upscaling}
\end{figure}

Conditional image generation networks are quite sensitive to the loss function used to evaluate reconstruction quality. Similarly to previous non-adversarial works, we use the VGG perceptual loss which extracts features from the predicted and actual images and computes the difference between them. It was found by \cite{Johnson2016Perceptual} to correlate with human perceptual similarity. A sketch of our method can be seen in Fig.~\ref{fig:upscaling}.

\subsection{Augmenting the input images}
\label{sec:method:augment}

In Sec.~\ref{sec:method:upscaling}, we proposed an upscaling network for conditional generation. By itself it is quite similar to previous conditional generation methods. The unique challenge here is the ability to generalize from a single image. We solve this task by extensive augmentations. Similarly to most other deep conditional generation works, we use crops and horizontal flips. However as just a single image is significantly less data then used in most other works, we use another non-linear augmentation, thin-plate-spline (TPS) \cite{bookstein1989principal}. Our TPS implementation proceeds in the following stages: i) it first constructs a target equi-spaced grid of size $4 \times 4$. We denote the target tile $t$. ii) it randomly transforms each grid point with magnitude determined by a scale factor iii) it learns a smooth TPS transformation $f$ with linear and radial terms which approximates the randomly warped target grid while preserving smoothness. The importance of each objective is parametrized by $\lambda$. The TPS loss objective is presented below:

\begin{equation}
    \label{eq:tps_loss}
    L_{TPS}(f) = \sum_{i, j} \|t(i, j) - f(i, j)\|^2 + \lambda \int \int f_{xx} + f_{yy} + 2f_{xy} dx dy
\end{equation}

This optimization can be performed very efficiently e.g. Donato and Belongie \cite{donato2002approximate}. The resulting transformation $f$ is then used to transform the original image $x$ for a training iteration. Different TPS warps are used for every training iteration.

\subsection{Learning a compact latent space}
\label{sec:method:vae}

The upsampling network proposed above can effectively upsample low-resolution images. This will be shown in Sec.~\ref{sec:exp} to be effective for conditional generation tasks. In this section, we propose a method for non-adversarial unconditional single image generation.

To perform unconditional generation, our method learns a compact latent space for the augmentations of the single input image. We propose to combine the upscaling network $G$, proposed in the previous sections with a variational autoencoder (VAE). The variational autoencoder consists of an encoder $E$ and decoder $D$. The encoder takes in a low resolution input image, which may be the original image or one of its augmentations - in both cases downsampled to the lowest resolution. The output of the encoder is a latent code of small spatial dimensionality and a larger number of channels (typically $3 \times 4 \times 128$). 

\begin{equation}
    \label{eq:encoder}
    z = E(x^0_{actual})
\end{equation}

We add normally distributed per-pixel random noise to the latent code, and pass it through the decoder network $D$:

\begin{equation}
    \label{eq:decoder}
    \tilde{x^0} = D(z + \epsilon)~~~\epsilon \sim N(0, I)
\end{equation}

We train the upscaling network and the VAE end-to-end, with the addition of a KL-divergence loss $L_{KL}(E) = \alpha \|z\|^2$ and a perceptual reconstruction loss on $x^{0}$, $L_0(E, D) = \ell(\tilde{x^0}, x^0_{actual})$. The full optimization loss is:

\begin{equation}
    \label{eq:vae}
    L_{total}(E, D, G) = L_{KL} + L_0 + L_{upscaling}
\end{equation}

\begin{figure}[t]
\begin{center}
\includegraphics[width=1\linewidth]{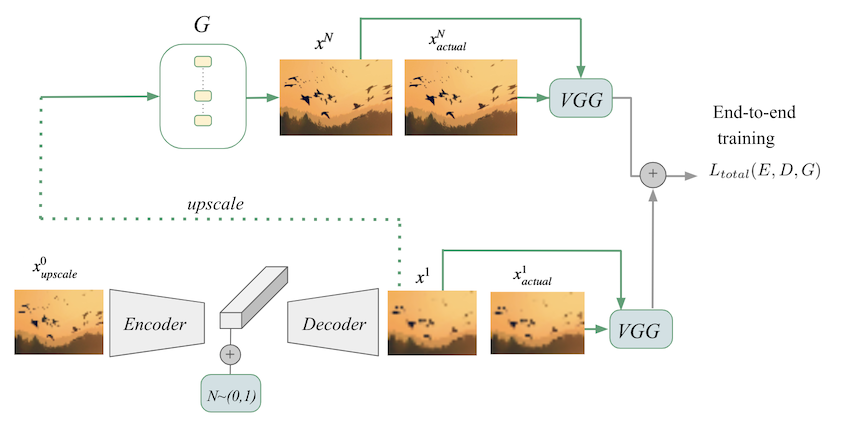}
\end{center}
\caption{An illustration of the variational autoencoder trained end-to-end with the upsampling network. The VAE consists of a low-resolution encoder and decoder, with corresponding KL loss on the latent space and perceptual reconstruction loss on the low-resolution input. }  
\label{fig:vae}
\end{figure}

\subsection{Synthesizing novel images with latent interpolations}
\label{sec:method:syn}

The objective for adding the VAE front end is to be able to perform image manipulations in latent space. Once the encoder, decoder and upsampling networks are trained, we can encode every input image $x$ into its latent code $z = E(x)$. For two different augmentations of the single input image: $x_1$ and $x_2$, we obtain the two codes $z_1$ and $z_2$. We can generate a novel image by interpolating between the two codes and generating a high-res image:

\begin{equation}
    \label{eq:inter}
    x_{novel} = G(D(\alpha \cdot z_1 + (1 - \alpha) \cdot z_2))
\end{equation}

Where $\alpha$ is a scalar between $(0,1)$. Furthermore, we can use the same idea for generating animations. To generate a short video clip from a single image (e.g. Fig.~\ref{fig:animation}), we encode two augmentations as described before and sample the interpolation with $\alpha$ sampled at regular intervals:

\begin{equation}
    \label{eq:inter_animate}
    x_{novel} = G(D(\alpha \cdot z_1 + (1 - \alpha) \cdot z_2)), \quad \alpha \in \bigcup{\left\{\frac{i}{T}\right\}}_{i=0}^{T}
\end{equation}

\subsection{Implementation details}
\label{sec:method:imp}
The VAE encoder is composed of $3$ convolutional blocks each followed by batch-norm and LeakyReLU. The convolutional blocks have $32, 64, 128$ channels respectively. The VAE decoder is the inverse of the encoder composed by deconvolutions and ReLU activations. At the end of the decoder, we put a Tanh activation. We do not learn noise $\sigma$ but rather used a constant value of $0.01$. The upscaling generator is a composition of $N$ (where $N$ depends on the resolution) blocks, each block consists of: $2$ convolutions, batch-norm, ReLU, another convolution, followed by an upsampling layer. The hyper-parameters follow the original implementation of SinGAN. In all of our experiments we trained with Adam optimizer with learning rate of $0.0005$,  $( \beta_1, \beta_2) = (0.5,0.999)$ and a cosine annealing schedule.

\section{Experiments}
\label{sec:exp}

In this section, we evaluate several applications highlighting the capabilities of our approach.

\subsection{Domain translation: Paint2Image, Edges2Image}

\begin{figure}[t]
\begin{center}
\begin{tabular}{@{\hskip0pt}c@{\hskip4pt}c@{\hskip2pt}c@{\hskip2pt}c@{\hskip2pt}c@{\hskip2pt}c@{\hskip2pt}c@{\hskip2pt}c}

Training Image & Input & SinGAN & Ours \\
\includegraphics[width=0.15\linewidth]{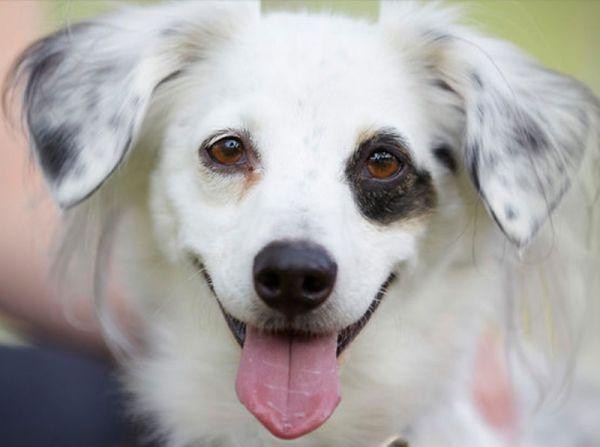} &
\includegraphics[width=0.15\linewidth]{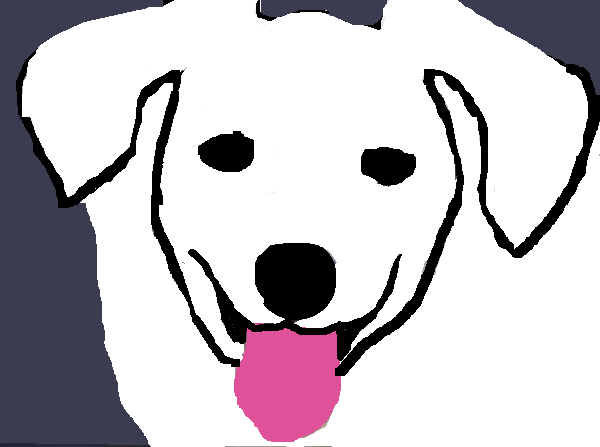} &
\includegraphics[width=0.15\linewidth]{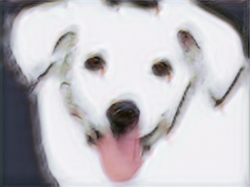} &
\includegraphics[width=0.15\linewidth]{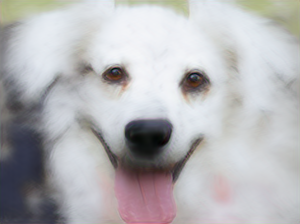} \\

\includegraphics[width=0.15\linewidth]{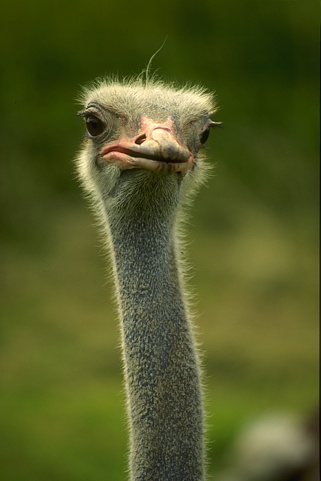} &
\includegraphics[width=0.15\linewidth]{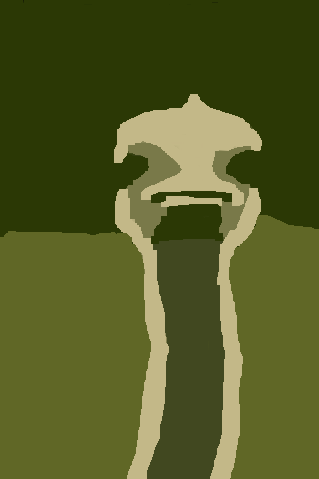} &
\includegraphics[width=0.15\linewidth]{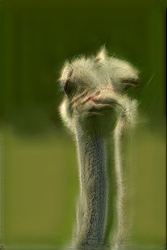} &
\includegraphics[width=0.15\linewidth]{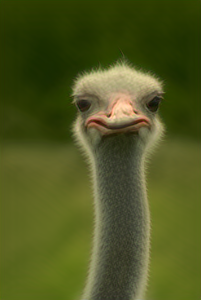} \\

\includegraphics[width=0.15\linewidth]{figures/paint2image/birds.png} &
\includegraphics[width=0.15\linewidth]{figures/paint2image/birds_paint_input.png} &
\includegraphics[width=0.15\linewidth]{figures/paint2image/SinGAN_Outputs/birds/birds_quant_start_scale=2.png} &
\includegraphics[width=0.15\linewidth]{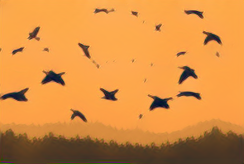} \\
\includegraphics[width=0.15\linewidth]{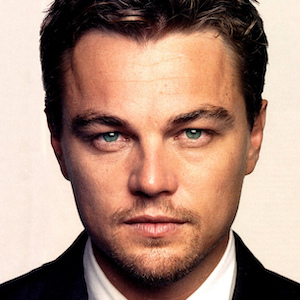} &
\includegraphics[width=0.15\linewidth]{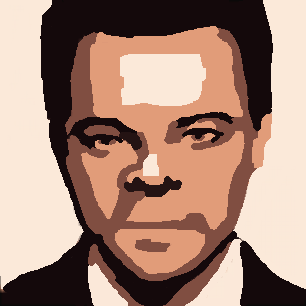} &
\includegraphics[width=0.15\linewidth]{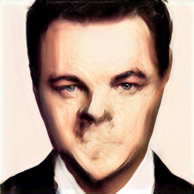} &
\includegraphics[width=0.15\linewidth]{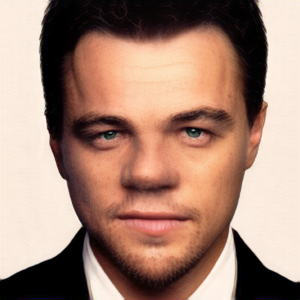} \\

\end{tabular}

\end{center}
\caption{Paint2Image: this figure presents our results on the painting to image task. (left to right) The training image, the painting used as input to the trained models, the results by SinGAN (with quantization), the results of our method. It is clear that our method results in more realistic images (SinGAN results are still more paint-like). It is remarkable that our method performs well on large objects such as human or animal faces whereas SinGAN does not perform as well in this setting.}
\label{fig:paint2image}
\end{figure}

We investigate the performance of our method on domain translation tasks. The task of painting to image, trains an upscaling network without the VAE front-end (as no latent manipulation is necessary). The input to the network is the training image after being downscaled to low-resolution and after color-quantization.  At inference time, we feed a downscaled painting into the upscaling network. The upscaled output can be observed in Fig.~\ref{fig:paint2image}. Our results are compared with SinGAN. Our method performs very well on such cases as can be seen in "Dog", "Ostrich" and "Face". SinGAN was unable to deal with larger objects, leaving them very similar to the original paint image. Our results are comparable with SinGAN on small objects such as "Birds". In Tab.~\ref{tab:SIFID} the methods are compared quantitatively in terms of Single Image FID \cite{shaham2019singan}. AugurOne outperforms SinGAN on this task for most images. Similarly, in the Edges2Image task, the input is a low-resolution edge image corresponding to the training image. The edge image can be obtained using a manual sketch or an automatic edge detector (we have experimented with both, and have found both options to work well). At test time a new sketch is provided to the network, the output is a novel photo corresponding to the sketch. Examples can be seen in Fig.~\ref{fig:edge2Image}, and many more can be seen in the appendix.

\begin{table}[b]
\centering
\begin{tabular}{lccccc}  
 & Dog  & Ostrich & Birds & Face & Average\\
 \midrule
SinGAN & 0.73& \textbf{0.40}& 0.44 & 1.17 & 0.68\\
AugurOne & \textbf{0.21} & \textbf{0.41} & \textbf{0.27} & \textbf{0.47} & \textbf{0.34}\\
 \bottomrule
\end{tabular}
\vspace{1em}
\caption{\label{tab:SIFID} Paint2Image: AugurOne has better higher than SinGAN (measured by SIFID)}
\end{table}

\begin{figure}[t]
\begin{center}
\includegraphics[width=1.0\linewidth]{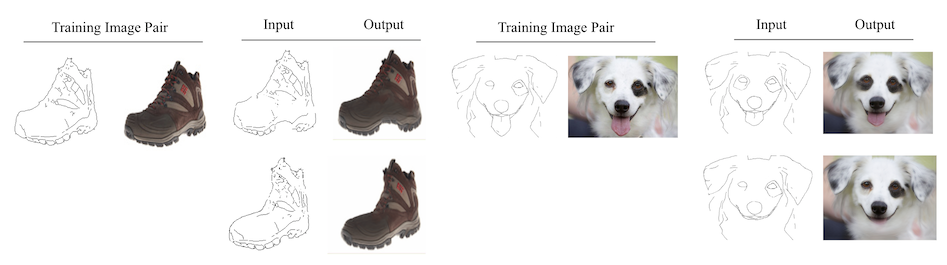} \\
\end{center}
\caption{Edges2Image: our method is trained on a pair of edges and photo. After training, we feed in a new edge map, our model synthesizes a new photo with the correct edges. Please note the change in shape of the shoes, and the movement of the dog's tongue.}
\label{fig:edge2Image}
\end{figure}

\subsection{Animating still images}

\begin{figure}[t]
\includegraphics[width=\textwidth]{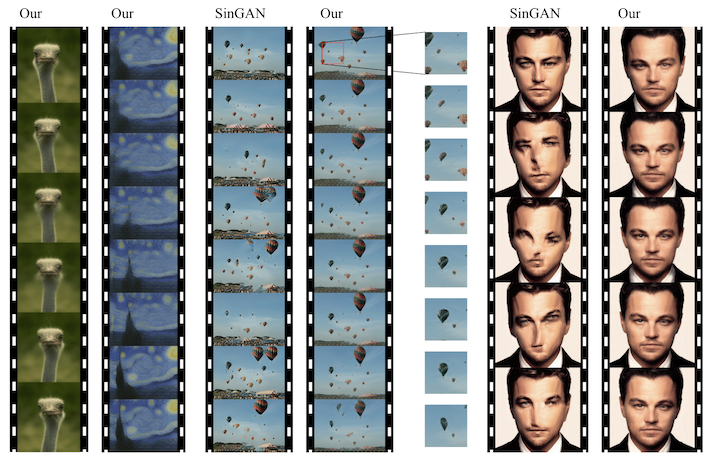}

\caption{Comparison of animations from a single image produced by SinGAN and our method. Our method produces animations by interpolation in latent space between two augmentations of the original image. We can observe from the "Balloons" animation that while SinGAN performs small local motions, our method can faithfully animate larger motions. The "Face" animation, demonstrates that while our method can animate large complex objects, SinGAN cannot handle such cases. We also present two other examples of animations of "Starry Night" and "Ostrich". \textit{We strongly urge the reader to view the animations in our \href{http://www.vision.huji.ac.il/augurone/}{\textcolor{blue}{project page}}, as the static version presented here does not fully capture the experience.}}
\label{fig:animation}
\end{figure}

Our method trains an encoder jointly with the generator. The learned encoder allows us to interpolate between two images and therefore generate compelling animations with large motions. \textit{To fully appreciate our results, we strongly urge the reader to view the videos on the \href{http://www.vision.huji.ac.il/augurone/}{\textcolor{blue}{project page}}.} Some examples of the results of our method can be observed in Fig.~\ref{fig:animation}. We can observe from the "Balloons" animation that our method generates more directed motions in comparison with SinGAN. The zoom-in column allows a better view of the motion synthesized by our method. Balloons are relatively small objects and feature texture-like images which SinGAN was designed for. We therefore also evaluate the two methods on face animation, a large complex object. In this case, our method is able to generate smooth and sharp animations, whereas SinGAN generates disfigured faces, demonstrating the superiority of our method on large structured objects. Our method is also evaluated on "Ostrich" and "Starry Night" and is shown to obtain pleasing animations.  

\subsection{Novel Images}

Our method can be extended to generate novel images of arbitrary size by the following procedure: i) concatenating an arbitrary number of  different random augmentations of the original image to form a pair of images of arbitrary size ii) Encode each of the pair of images into latent codes. iii) Interpolate between the two random latent codes and project the interpolated latent code through the generator to form a high-resolution image.

\begin{figure}[t]
\includegraphics[width=\textwidth]{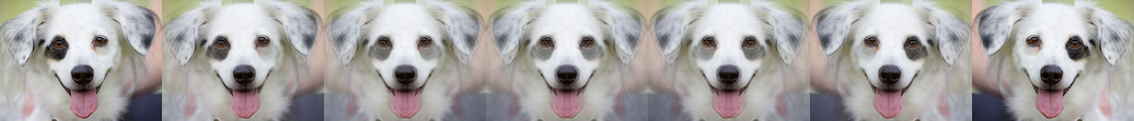}\\
\includegraphics[width=\textwidth]{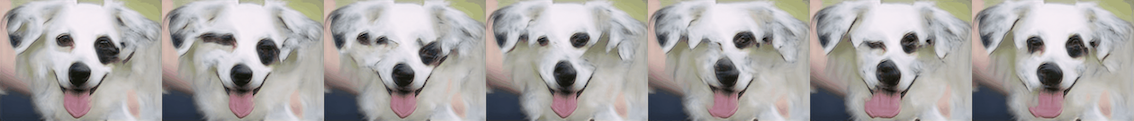}\\
\includegraphics[width=\textwidth]{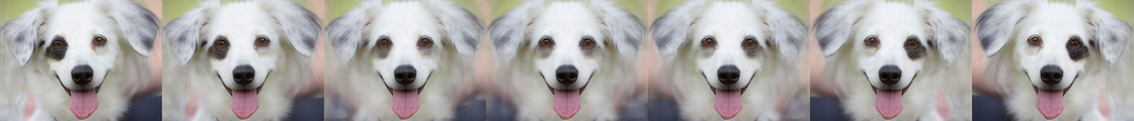}\\
\includegraphics[width=\textwidth]{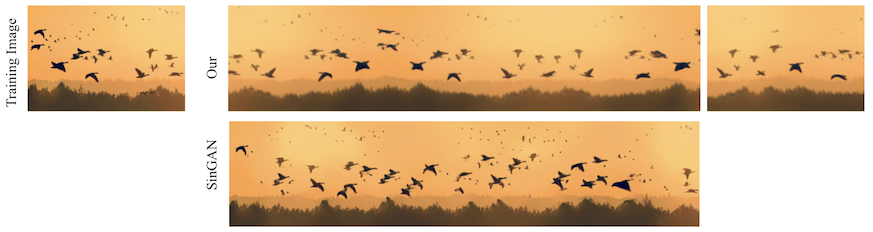}
\caption{(top to bottom) Random Dog samples generated by interpolating between two augmentations using (i) linear interpolation (ii) SinGAN (iii) our method. Linear interpolation and SinGAN do not generate realistic intermediate images. Our methods generates realistic novel images. (iv) Novel image synthesis with our method and (iv) SinGAN. The two methods perform comparably on the "Birds" image.  }
\label{fig:linear_interpolation}

\end{figure}

Several examples of our results can be observed in Fig.~\ref{fig:linear_interpolation}. We can see from "Dog" examples that our method is able to generate compeling novel images of the dog. SinGAN fails on this image, as it struggles to deal with large complex objects. We can see that simple techniques such as linear interpolation do not generate compelling novel images. We also presented an example of generating "Birds" images of arbitrary size. Our method performs comparably to SinGAN on this task. As a failure mode of our method, we should note that highly textured scenes are sometimes imperfectly captured by the latent space of the VAE, and our results can result in blurrier results than SinGAN in such cases.

\subsection{Image Harmonization}

\begin{figure}[t]
\begin{center}
\begin{tabular}{@{\hskip0pt}c@{\hskip4pt}c@{\hskip2pt}c@{\hskip2pt}c@{\hskip2pt}c@{\hskip2pt}c@{\hskip2pt}c@{\hskip2pt}c}

Training Image & Input & SinGAN & Ours \\
\includegraphics[width=0.25\linewidth]{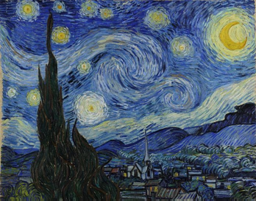} &
\includegraphics[width=0.25\linewidth]{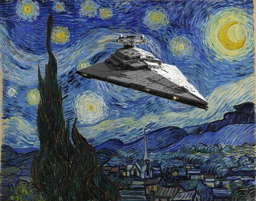} &
\includegraphics[width=0.25\linewidth]{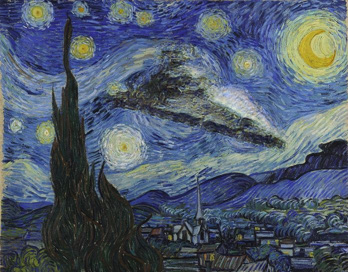} &
\includegraphics[width=0.25\linewidth]{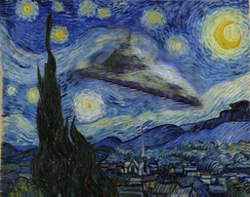} \\

\includegraphics[width=0.25\linewidth]{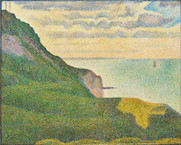} &
\includegraphics[width=0.25\linewidth]{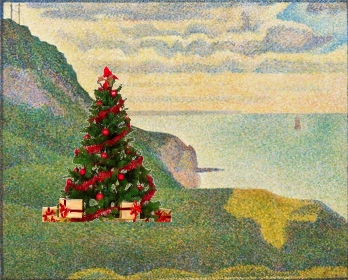} &
\includegraphics[width=0.25\linewidth]{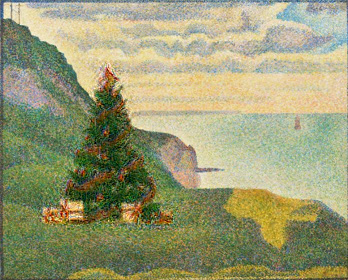} &
\includegraphics[width=0.25\linewidth]{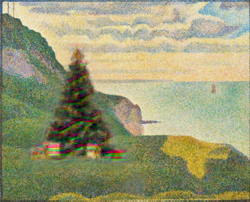} \\

\end{tabular}

\end{center}
\caption{Image harmonization experiments: (left to right) Training image, the input composition of the foreground object with the original image as background, output of SinGAN, output of our method. Both SinGAN and our method use the foreground segmentation mask. Our method blends the foreground object naturally into the input image.}  
\label{fig:harmonization}
\vspace{1em}
\end{figure}

Our method is evaluated on image harmonization in Fig.~\ref{fig:harmonization}. The task of image harmonization places an external object on top of the single image that the generator was trained on. The task is to seamlessly blend the external object with the background image. We train our upscaling network without the VAE front-end (as there is no need for latent interpolation for this task). Similarly to SinGAN, we perform harmonization using injection at an intermediate block of the network. It can be observed that our method generates harmonious images.

\section{Discussion}
\label{sec:disc}

\textbf{The motivation for using a single image:} Generative models are typically trained on large image collections and have achieved amazing results, in particular on face image generation. In this work we tackled the challenging task of training generative models from a single image. Our motivation for training models from a single image, is that for the long-tail of images we will not have image collections that are very close to it. Such collections only exist in for objects of wide interest (e.g. faces) or for very diverse datasets that require models to generalize to modeling many different types of images, an important but difficult task. Instead, single image generation makes no assumptions on the availability of large collection. Another motivation is the cost of obtaining copy-rights to a large number of images, a cost that may be prohibitive. Instead, simply training based on a single image can be performed by the image owner with no additional cost.

\textbf{Pre-trained perceptual loss:} In this work, we relied strongly on the availability of pre-trained deep perceptual losses. It is sometimes argued that this amounts to using significant supervision as such perceptual losses are trained on very large supervised datasets (in our case, on the ImageNet dataset). We argue that this does not suffer from any of the disadvantages that we highlighted for large dataset generative models as imagenet-pretrained models are easily available at no cost and require no extra supervision for new tasks such as single-image generator training.

\textbf{Non-adversarial Learning:} Generative adversarial networks have dominated image generation and translation over the last few years. Although they have many well documented advantages, they have disadvantages in terms of stability and mode collapse. We suspect that GANs have been used for applications that do not require them. This is further motivated by previous work on non-adversarial learning e.g. image generation, unsupervised domain translation. In this work, we showed that our non-adversarial approach has yielded a simpler method which could be trained end-to-end. Furthermore, the modeling flexibility has enabled us to include an encoder and be able to deal with whole objects rather than texture only. This enables us to generate object-level single image animations. One advantage of GAN methods over our method is their good performance as a perceptual loss for textures, which we observed to be better than non-adversarial perceptual losses.   


\textbf{Super-resolution:} Our method can be trained to perform super-resolution in a very similar manner to ZSSR \cite{shocher2018zero}. Indeed, our upsampling network with a single stage is not very different from the network used by ZSSR. Similarly to this method, we can obtain more faithful reconstruction, while suffering a little in terms of perceptual quality than GAN methods. This lies on a different point of the perception-distortion curve than SinGAN.

\section{Conclusions}
\label{sec:conc}

We analysed a recent method for training image generators from a single image. Our analysis simplified SinGAN into 3 parts: upsampling, augmentation and latent manipulation. This lead to the development of a novel non-adversarial approach for single image generative modeling. Our approach was shown to handle operations on large objects better than previous methods, generating compelling single-image animations with large motion and performing effective translation between domains.   

\clearpage
%
%
\bibliographystyle{splncs04}
\bibliography{egbib}

\section{Appendix}
\label{sec:appendix}
\subsection{Additional Edges2Image Results}
\label{sec:app:edge2image}
\begin{figure}[t]
\begin{center}
\begin{tabular}{@{\hskip2pt}c@{\hskip2pt}c@{\hskip2pt}c@{\hskip2pt}c}

Training Edge & Training Image & Input Edge & Output Image \\
\midrule
\multicolumn{4}{c}{Reducing the mouth size and cutting the chin} \\
\midrule 
\\
\includegraphics[width=0.25\linewidth]{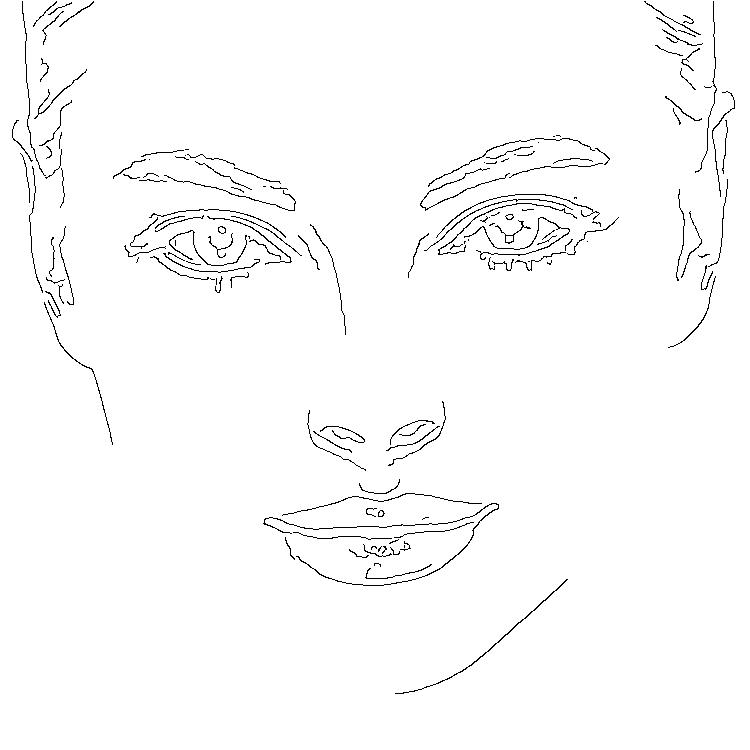} &
\includegraphics[width=0.25\linewidth]{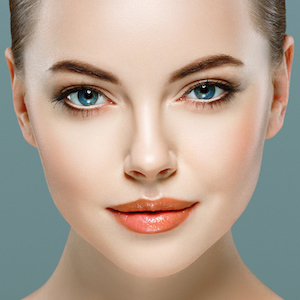} &
\includegraphics[width=0.25\linewidth]{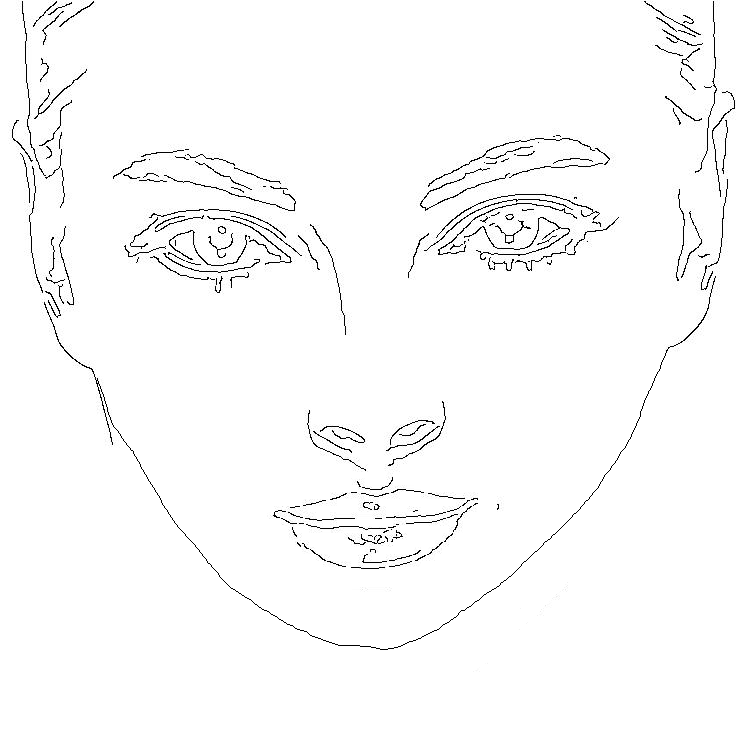} &
\includegraphics[width=0.25\linewidth]{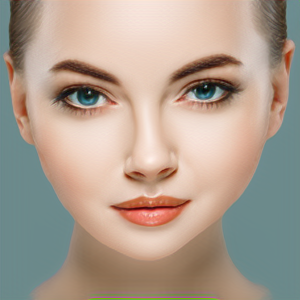}\\
\midrule
\multicolumn{4}{c}{Reducing the size of the eyes and stretching the mouth} \\
\midrule 
\\
\includegraphics[width=0.25\linewidth]{figures/sm/edges2im/face/face_a_edge.jpg} &
\includegraphics[width=0.25\linewidth]{figures/sm/edges2im/face/face_a.jpg} &
\includegraphics[width=0.25\linewidth]{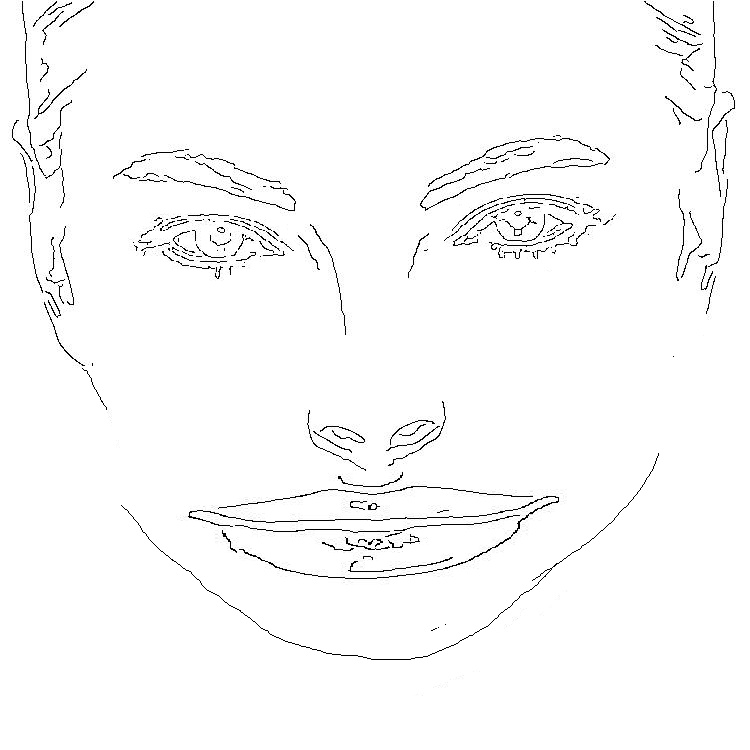} &
\includegraphics[width=0.25\linewidth]{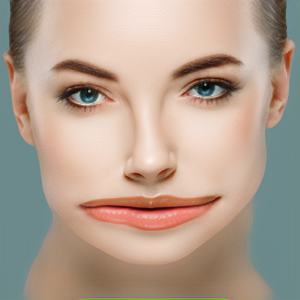}\\
\midrule
\multicolumn{4}{c}{Lifting the nose} \\
\midrule 
\\
\includegraphics[width=0.25\linewidth]{figures/sm/edges2im/face/face_a_edge.jpg} &
\includegraphics[width=0.25\linewidth]{figures/sm/edges2im/face/face_a.jpg} &
\includegraphics[width=0.25\linewidth]{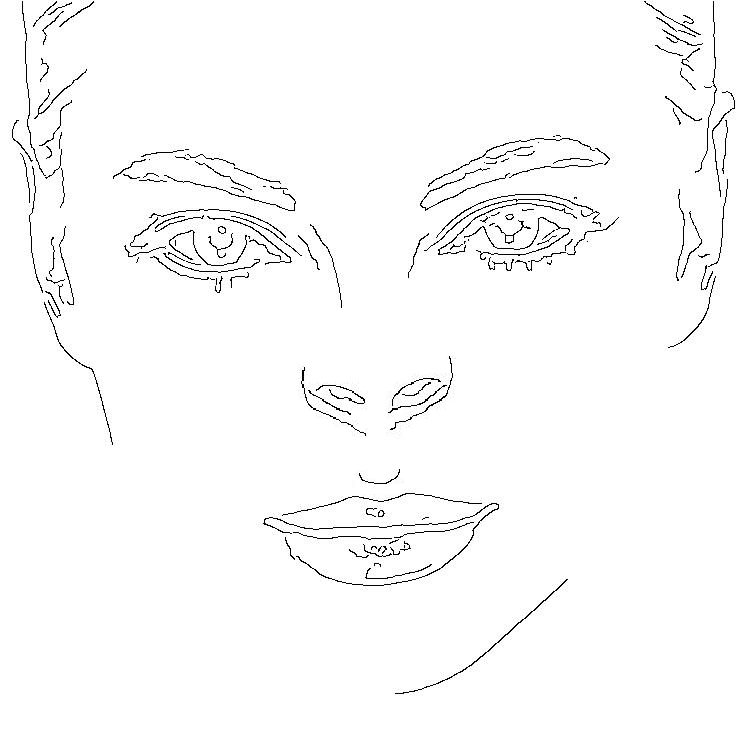} &
\includegraphics[width=0.25\linewidth]{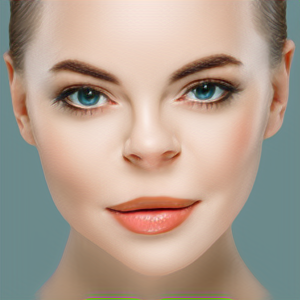}\\

\end{tabular}

\end{center}
\caption{Edges2Image examples }
\label{fig:app:edge2image:face}
\end{figure}

\begin{figure}[t]
\begin{center}
\begin{tabular}{@{\hskip2pt}c@{\hskip2pt}c@{\hskip2pt}c@{\hskip2pt}c}

Training Edge & Training Image & Input Edge & Output Image \\
\midrule
\multicolumn{4}{c}{Adding a large balloon} \\
\midrule 
\\
\includegraphics[width=0.25\linewidth]{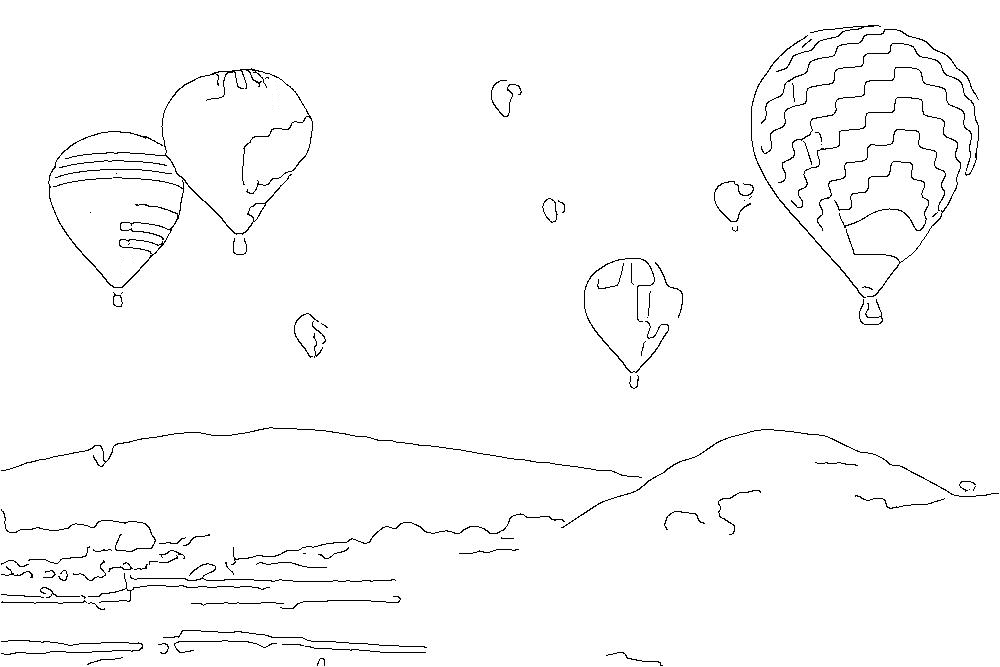} &
\includegraphics[width=0.25\linewidth]{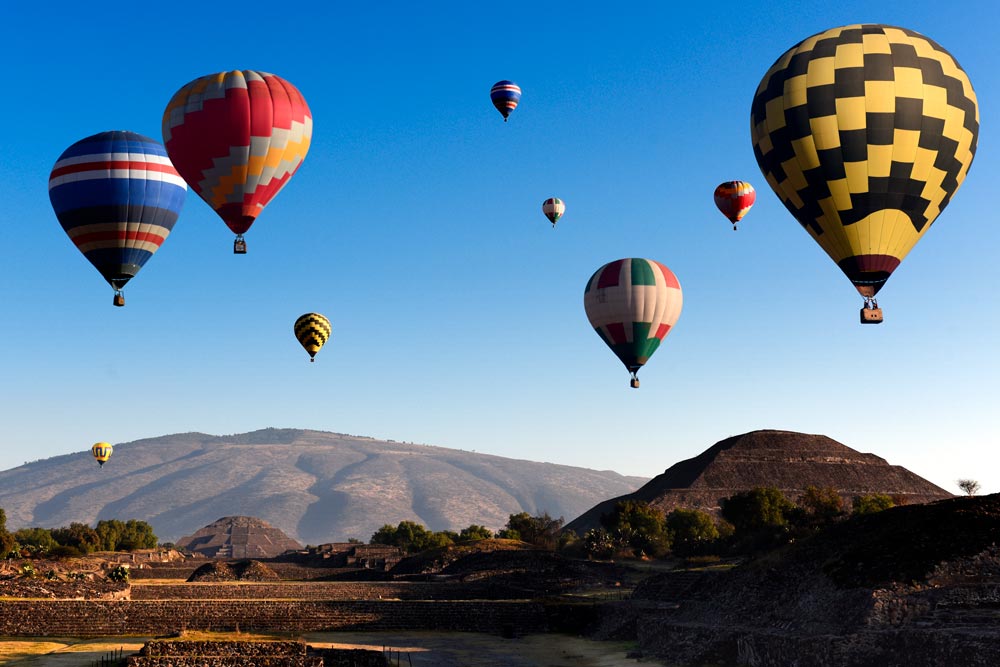} &
\includegraphics[width=0.25\linewidth]{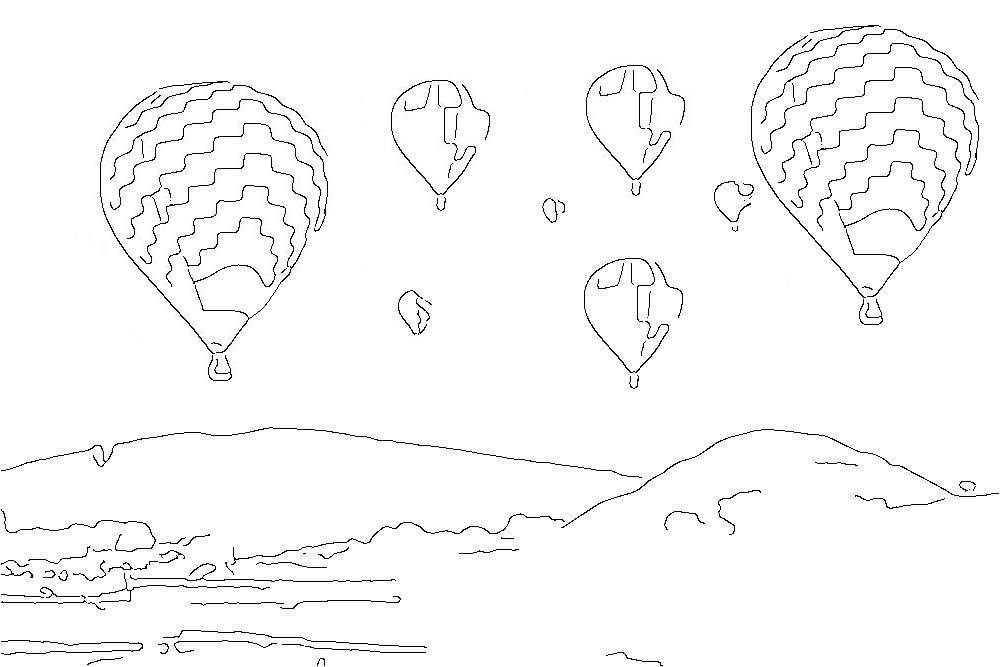} &
\includegraphics[width=0.25\linewidth]{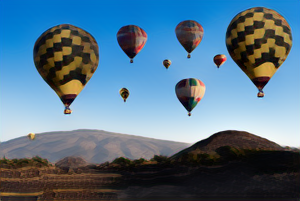}
\\
\midrule
\multicolumn{4}{c}{Adding distant mountains} \\
\midrule 
\\
\includegraphics[width=0.25\linewidth]{figures/sm/edges2im/baloons/baloons_edge.png} &
\includegraphics[width=0.25\linewidth]{figures/sm/edges2im/baloons/baloons.jpg} &
\includegraphics[width=0.25\linewidth]{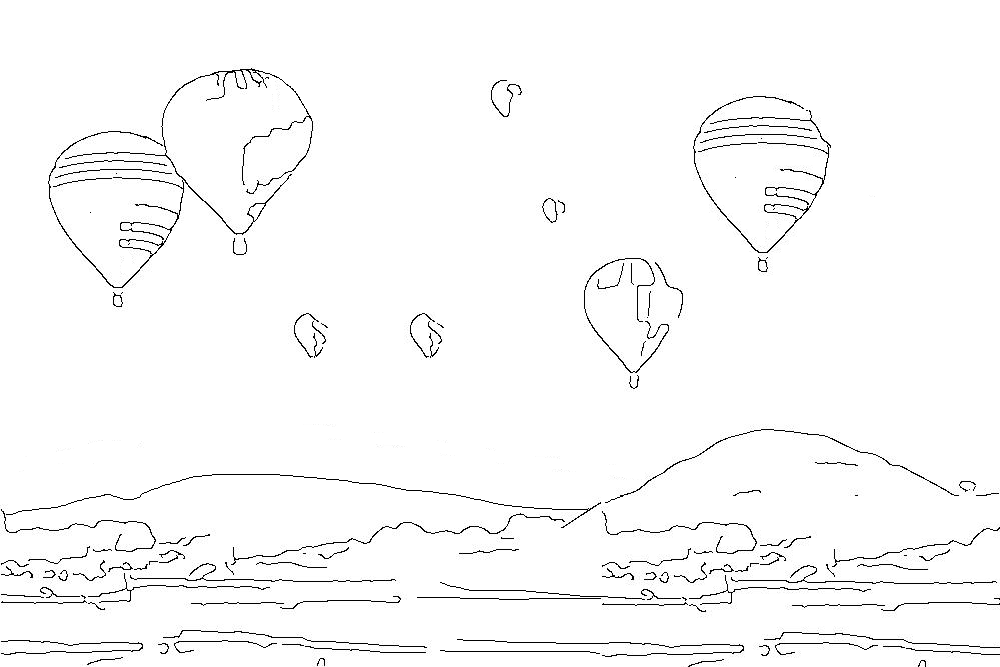} &
\includegraphics[width=0.25\linewidth]{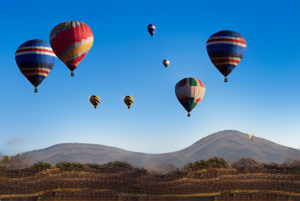}\\
\midrule
\multicolumn{4}{c}{Lowering hand of Kuala} \\
\midrule 
\\
\includegraphics[width=0.25\linewidth]{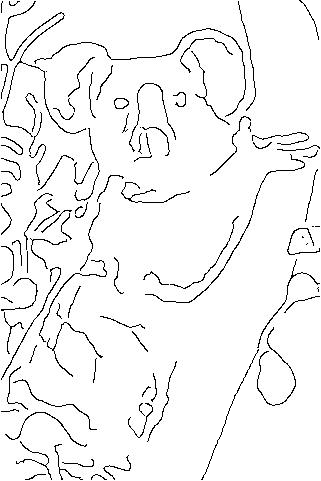} &
\includegraphics[width=0.25\linewidth]{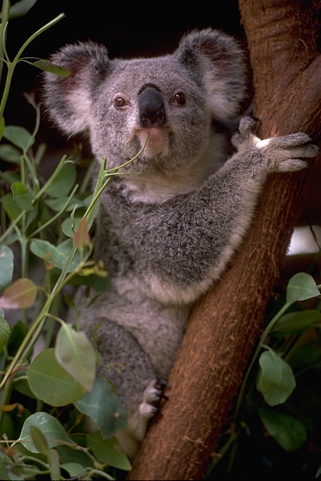} &
\includegraphics[width=0.25\linewidth]{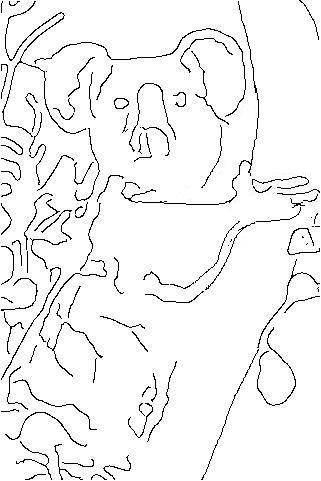}&
\includegraphics[width=0.25\linewidth]{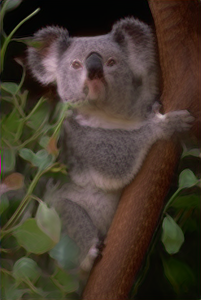} \\
\midrule
\multicolumn{4}{c}{Moving the tree} \\
\midrule 
\\
\includegraphics[width=0.25\linewidth]{figures/sm/edges2im/kuala/original_image_edge.png} &
\includegraphics[width=0.25\linewidth]{figures/sm/edges2im/kuala/original_image.jpg} &
\includegraphics[width=0.25\linewidth]{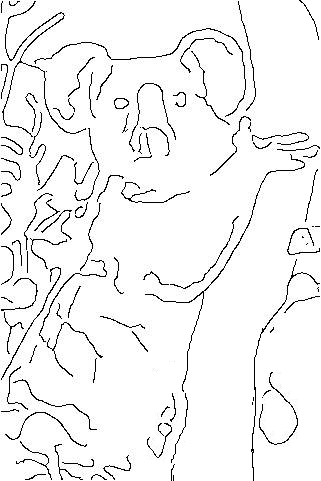}&
\includegraphics[width=0.25\linewidth]{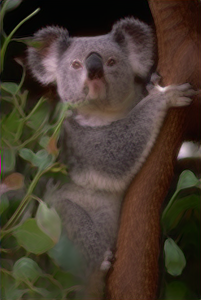}\\
\end{tabular}

\end{center}
\caption{Edges2Image examples }
\label{fig:app:edge2image:baloons}
\end{figure}

\begin{figure}[t]
\begin{center}
\begin{tabular}{cccc}

Training Edge & Training Image & Input Edge & Output Image \\
\midrule
\multicolumn{4}{c}{Changing buckle, narrowing shoe} \\
\midrule 
\\
\includegraphics[width=0.25\linewidth]{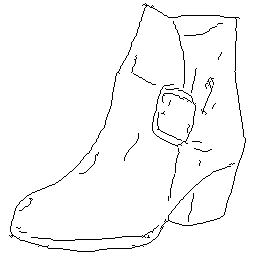} &
\includegraphics[width=0.25\linewidth]{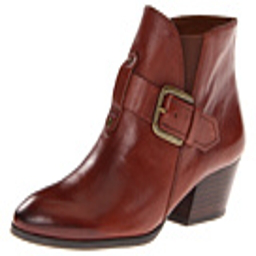} &
\includegraphics[width=0.25\linewidth]{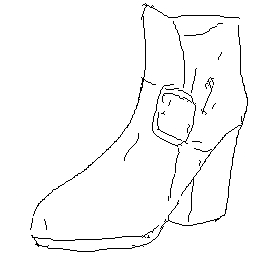}&
\includegraphics[width=0.25\linewidth]{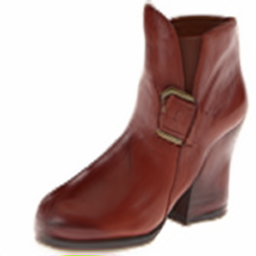} \\
\midrule
\multicolumn{4}{c}{Lowering the buckle} \\
\midrule 
\\
\includegraphics[width=0.25\linewidth]{figures/sm/edges2im/y_012/original_im_edge.png} &
\includegraphics[width=0.25\linewidth]{figures/sm/edges2im/y_012/original_im.png}&
\includegraphics[width=0.25\linewidth]{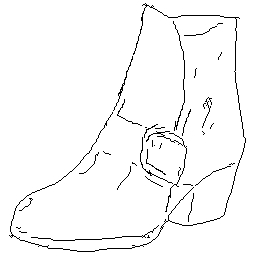}&
\includegraphics[width=0.25\linewidth]{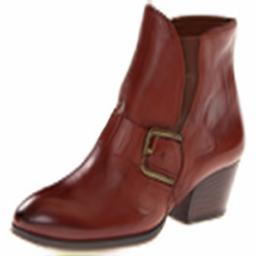}\\
\midrule
\multicolumn{4}{c}{Shortening the upper part} \\
\midrule 
\\
\includegraphics[width=0.25\linewidth]{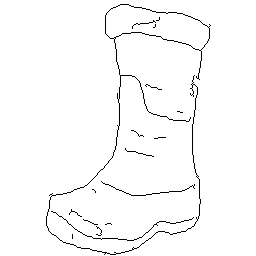} &
\includegraphics[width=0.25\linewidth]{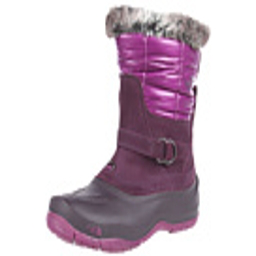} &
\includegraphics[width=0.25\linewidth]{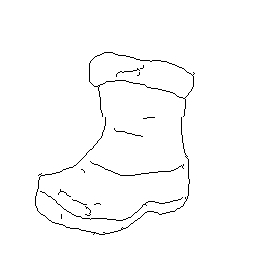}&
\includegraphics[width=0.25\linewidth]{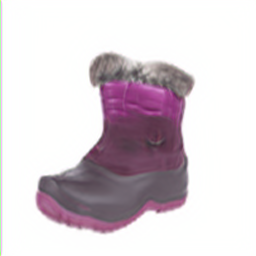} \\
\midrule
\multicolumn{4}{c}{Increasing height of heel} \\
\midrule 
\\
\includegraphics[width=0.25\linewidth]{figures/sm/edges2im/y_069/y_069_edge.png} &
\includegraphics[width=0.25\linewidth]{figures/sm/edges2im/y_069/y_069_im.png} &
\includegraphics[width=0.25\linewidth]{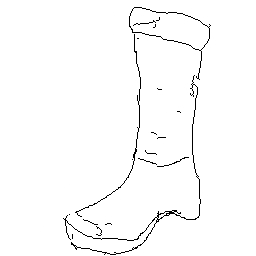}&
\includegraphics[width=0.25\linewidth]{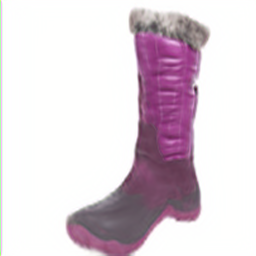}

\end{tabular}

\end{center}
\caption{Edges2Image examples}
\label{fig:app:edge2image:shoes}
\end{figure}

\begin{figure}[t]
\begin{center}
\begin{tabular}{cccc}

Training Edge & Training Image & Input Edge & Output Image \\
\midrule
\multicolumn{4}{c}{Adding a leg} \\
\midrule 
\\
\includegraphics[width=0.25\linewidth]{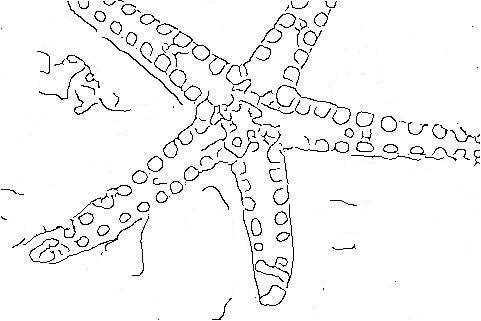} &
\includegraphics[width=0.25\linewidth]{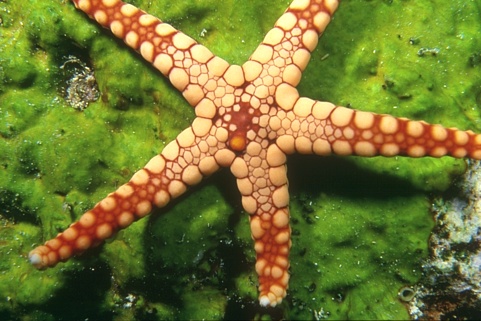} &
\includegraphics[width=0.25\linewidth]{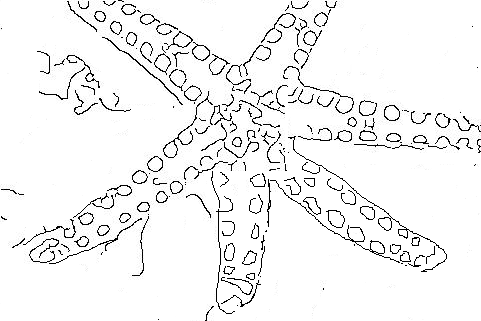}&
\includegraphics[width=0.25\linewidth]{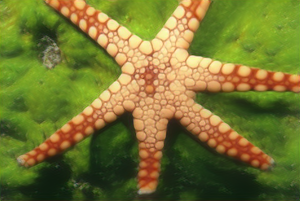}\\
\midrule
\multicolumn{4}{c}{Transforming into a snake} \\
\midrule 
\\
\includegraphics[width=0.25\linewidth]{figures/sm/edges2im/star/original_image_edge.png} &
\includegraphics[width=0.25\linewidth]{figures/sm/edges2im/star/original_image.jpg} &
\includegraphics[width=0.25\linewidth]{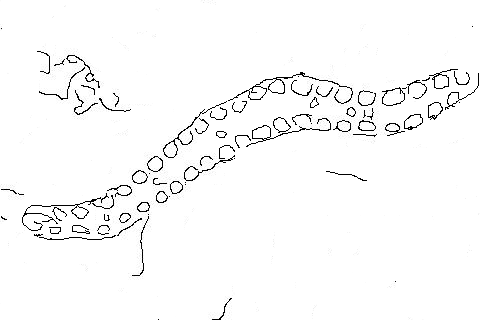}&
\includegraphics[width=0.25\linewidth]{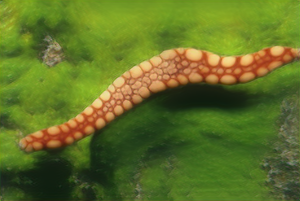}

\end{tabular}

\end{center}
\caption{Edges2Image examples}
\label{fig:app:edge2image:star}
\end{figure}

We present additional examples of our Edge2Image results. They were obtained by: i) taking an input hi-res image ii) computing its edges using the Canny edge detector iii) training our upscaling network, taking as input a downscaled edge image and the full hi-res image, the network is trained to predict the hi-res image using the low-res edges iv) augment the single image and edge pair by crops, flips and most importantly thin plate spline (TPS) transformations. Inference was performed by drawing a new edge image and inputting into our trained network (after downscaling). Results are presented in Figs.~\ref{fig:app:edge2image:face}~-~\ref{fig:app:edge2image:star}. Our method is able to perform significant image manipulations resulting in very high quality outputs. 

\subsection{Additional Paint2Image Results}
\label{sec:app:paint2image}

\begin{figure}[t]
\begin{center}
\begin{tabular}{@{\hskip2pt}c@{\hskip2pt}c@{\hskip2pt}c}

Training Image & Input Paint & Output Image \\
\midrule
\multicolumn{3}{c}{Stretching the cheeks} \\
\midrule 
\\
\includegraphics[width=0.3\linewidth]{figures/paint2image/face.png} &
\includegraphics[width=0.3\linewidth]{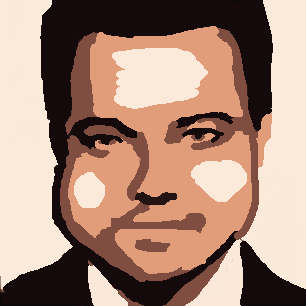} &
\includegraphics[width=0.3\linewidth]{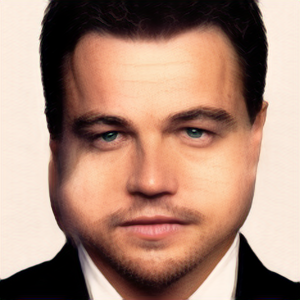}
\\
\midrule
\multicolumn{3}{c}{Lifting and stretching the nose} \\

\midrule 
\\
\includegraphics[width=0.3\linewidth]{figures/paint2image/face.png} &
\includegraphics[width=0.3\linewidth]{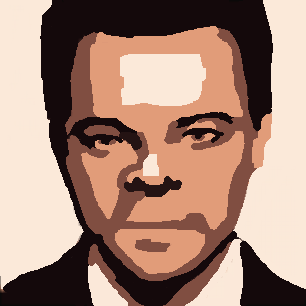} &
\includegraphics[width=0.3\linewidth]{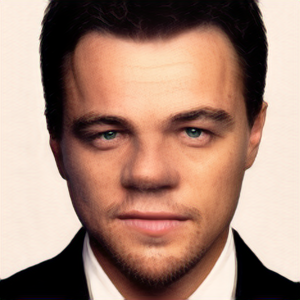}
\\
\midrule
\multicolumn{3}{c}{Shrinking the chin and the mouth} \\
\midrule 
\\
\includegraphics[width=0.3\linewidth]{figures/paint2image/face.png} &
\includegraphics[width=0.3\linewidth]{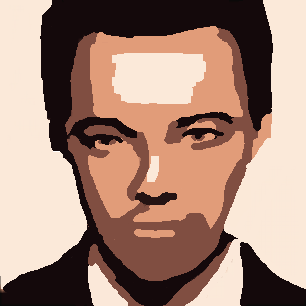} &
\includegraphics[width=0.3\linewidth]{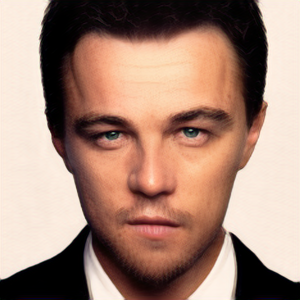}

\end{tabular}

\end{center}
\caption{Paint2Image: Our method is able to generalize from a single input image to highly variable input paintings. }
\label{fig:app:paint2image:face}
\end{figure}

\begin{figure}[t]
\begin{center}
\begin{tabular}{@{\hskip2pt}c@{\hskip2pt}c@{\hskip2pt}c}
Training Image & Input Paint & Output Image \\
\midrule
\multicolumn{3}{c}{Changing the position of the head and the neck, increasing the forehead} \\
\midrule 
\\
\includegraphics[width=0.26\linewidth]{figures/paint2image/ostrich_training_image.jpg} &
\includegraphics[width=0.26\linewidth]{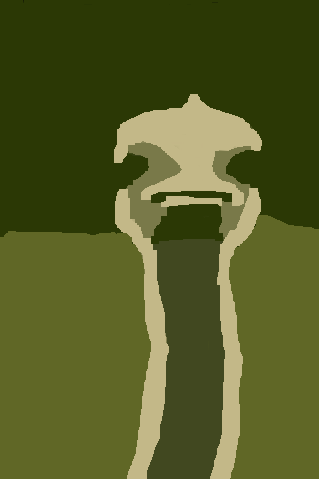} &
\includegraphics[width=0.26\linewidth]{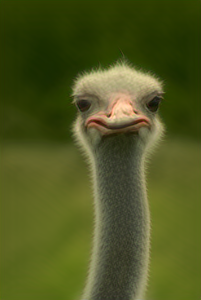}
\\
\midrule
\multicolumn{3}{c}{Changing the neck's position and shape} \\
\midrule 
\\
\includegraphics[width=0.26\linewidth]{figures/paint2image/ostrich_training_image.jpg} &
\includegraphics[width=0.26\linewidth]{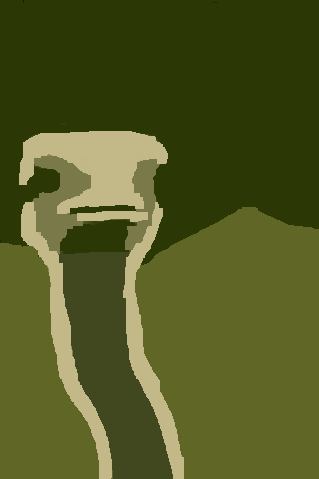} &
\includegraphics[width=0.26\linewidth]{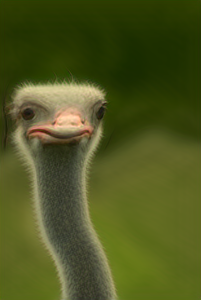}
\\
\midrule
\multicolumn{3}{c}{Changing point of view} \\
\midrule 
\\
\includegraphics[width=0.26\linewidth]{figures/paint2image/ostrich_training_image.jpg} &
\includegraphics[width=0.26\linewidth]{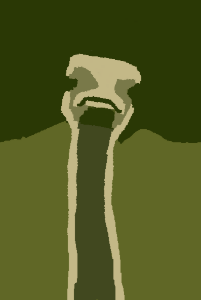} &
\includegraphics[width=0.26\linewidth]{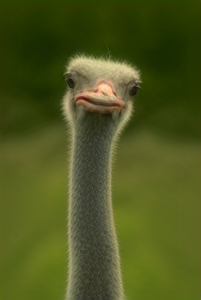}

\end{tabular}

\end{center}
\caption{Paint2Image on the "Ostrich" image: Our method is able to generalize from a single input image to highly variable input paintings. }
\label{fig:app:paint2image:ostrich}
\end{figure}

\begin{figure}[t]
\begin{center}
\begin{tabular}{@{\hskip2pt}c@{\hskip2pt}c@{\hskip2pt}c}
Training Image & Input Paint & Output Image \\
\midrule
\multicolumn{3}{c}{Stretching the nose to the right} \\
\midrule 
\\
\includegraphics[width=0.3\linewidth]{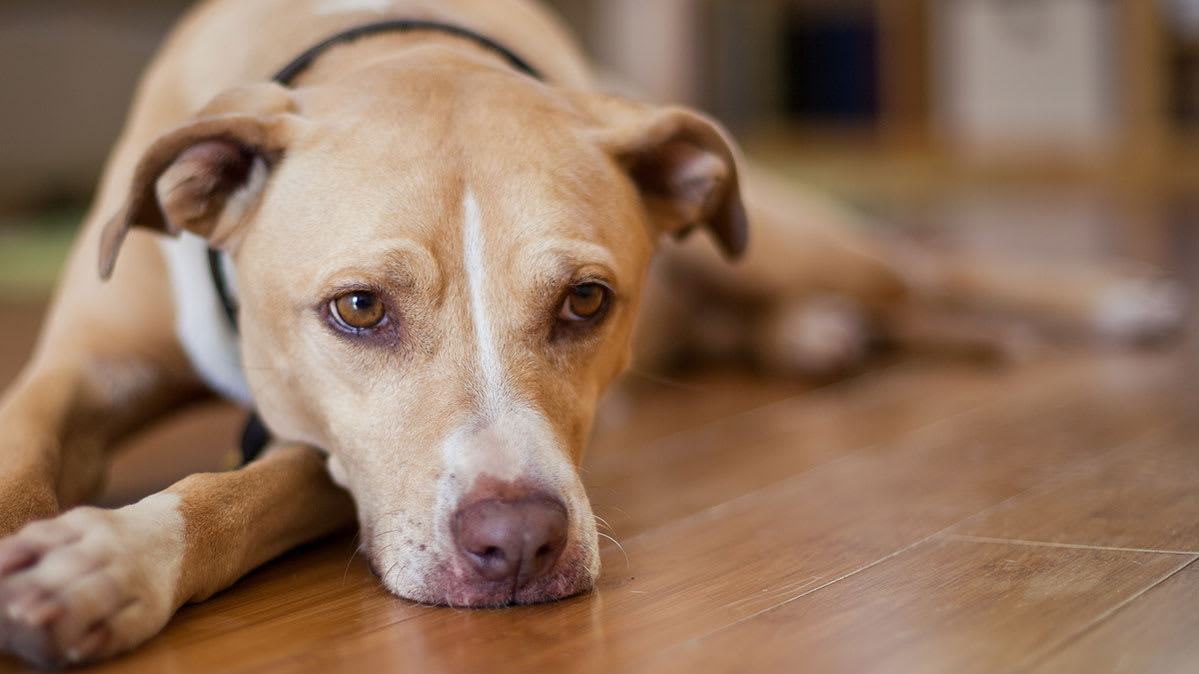} &
\includegraphics[width=0.3\linewidth]{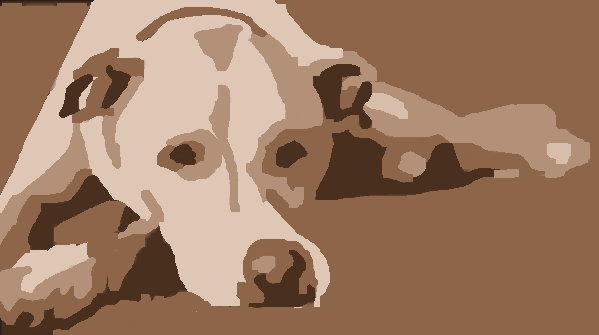} &
\includegraphics[width=0.3\linewidth]{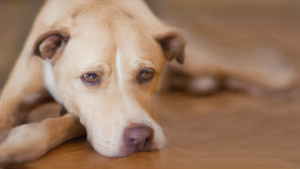}
\\
\midrule
\multicolumn{3}{c}{Stretching the nose to the middle} \\
\midrule 
\\
\includegraphics[width=0.3\linewidth]{figures/sm/paint2image/dog2/dog2.jpeg} &
\includegraphics[width=0.3\linewidth]{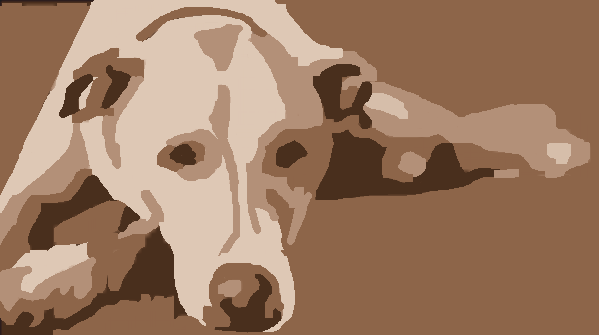} &
\includegraphics[width=0.3\linewidth]{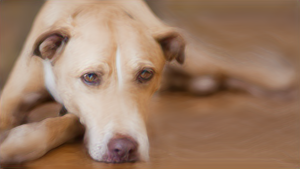}
\\
\midrule
\multicolumn{3}{c}{Shrinking the nose} \\
\midrule 
\\
\includegraphics[width=0.3\linewidth]{figures/sm/paint2image/dog2/dog2.jpeg} &
\includegraphics[width=0.3\linewidth]{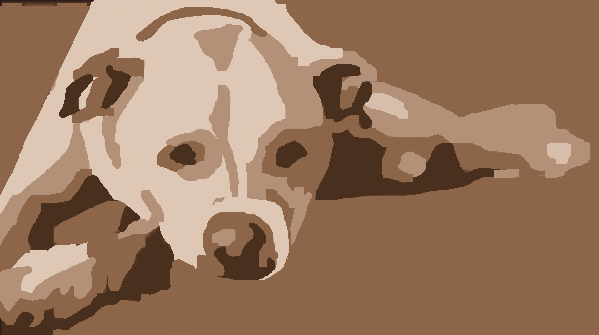} &
\includegraphics[width=0.3\linewidth]{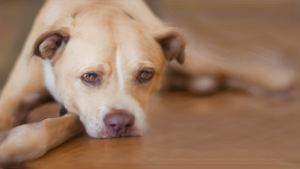}

\end{tabular}

\end{center}
\caption{Paint2Image on the "Dog" image: Our method is able to generalize from a single input image to highly variable input paintings. }
\label{fig:app:paint2image:dog}
\end{figure}

We present more examples of our Paint2Image results. They were obtained by standard training of our upsampilng network. Input low-res images were quantized during training (as done in SinGAN). Results are shown in Figs.~\ref{fig:app:paint2image:face}~-~\ref{fig:app:paint2image:dog}. Our method is able to obtain extremely high-quality results generalizing from paint to high-quality image outputs.

\end{document}